\documentclass[letterpaper]{article} 
\usepackage{aaai23}  
\usepackage{times}  
\usepackage{helvet}  
\usepackage{courier}  
\usepackage[hyphens]{url}  
\usepackage{graphicx} 
\usepackage{natbib}  
\usepackage{caption} 
\usepackage{algorithm}
\usepackage{algorithmic}
\usepackage{amsmath}
\usepackage{fontawesome}
\usepackage{array}
\usepackage{amssymb}
\usepackage{newfloat}
\usepackage{listings}
\DeclareCaptionStyle{ruled}{labelfont=normalfont,labelsep=colon,strut=off} 
\floatstyle{ruled}
\newfloat{listing}{tb}{lst}{}
\floatname{listing}{Listing}
\title{GPTR: Gestalt-Perception Transformer for Diagram Object Detection}
\author {
	Xin Hu,\textsuperscript{\rm 1,\rm 2}
	Lingling Zhang, \textsuperscript{\rm 1,\rm 2}\thanks{Corresponding author.}
	Jun Liu \textsuperscript{\rm 1,\rm 2}
	Jinfu Fan,\textsuperscript{\rm 3}
	Yang You, \textsuperscript{\rm 4}
	Yaqiang Wu \textsuperscript{\rm 2,\rm 5}
}
\affiliations {
	\textsuperscript{\rm 1} Shaanxi Provincial Key Laboratory of Big Data Knowledge Engineering, \\School of Computer Science and Technology, Xi’an Jiaotong University, China\\
	\textsuperscript{\rm 2} National Engineering Lab for Big Data Analytics, Xi'an Jiaotong University, China\\
	\textsuperscript{\rm 3} Department of Control Science and Engineering, Tongji University, Shanghai, China\\
	\textsuperscript{\rm 4} Department of Computer Science, National University of Singapore, Singapore\\
	\textsuperscript{\rm 5} Lenovo Research, Beijing, China\\
	dr.huxin711@foxmail.com, \{zhanglling, liukeen\}@xjtu.edu.cn, 1910648@tongji.edu.cn, youy@comp.nus.edu.sg, wuyqe@lenovo.com
}

\usepackage{bibentry}

\begin{document}

\maketitle

\begin{abstract}
Diagram object detection is the key basis of practical applications such as textbook question answering. Because the diagram mainly consists of simple lines and color blocks, its visual features are sparser than those of natural images. In addition, diagrams usually express diverse knowledge, in which there are many low-frequency object categories in diagrams. These lead to the fact that traditional data-driven detection model is not suitable for diagrams. In this work, we propose a gestalt-perception transformer model for diagram object detection, which is based on an encoder-decoder architecture. Gestalt perception contains a series of laws to explain human perception, that the human visual system tends to perceive patches in an image that are similar, close or connected without abrupt directional changes as a perceptual whole object. Inspired by these thoughts, we build a gestalt-perception graph in transformer encoder, which is composed of diagram patches as nodes and the relationships between patches as edges. This graph aims to group these patches into objects via laws of similarity, proximity, and smoothness implied in these edges, so that the meaningful objects can be effectively detected. The experimental results demonstrate that the proposed GPTR achieves the best results in the diagram object detection task. Our model also obtains comparable results over the competitors in natural image object detection.
\end{abstract}

\section{Introduction}
The goal of object detection \cite{liu2020deep,guo2021distilling,dong2021bridging} is to accurately locate and classify all objects in a given image, which is indeed dominated by various deep neural networks \cite{pan2021model,wu2021generalized,zhong2021dap,chen2021scale,cao2021few,wang2021end}. For this task, it is very important to understand the detailed and implicit semantic information of images. It also has great significance in practical applications such as visual question answering \cite{yuan2021perception}, cross-modal retrieval \cite{chen2021learning,diao2021similarity}, etc.

\begin{figure}[t]
	\centering
	\includegraphics[width=0.44\textwidth]{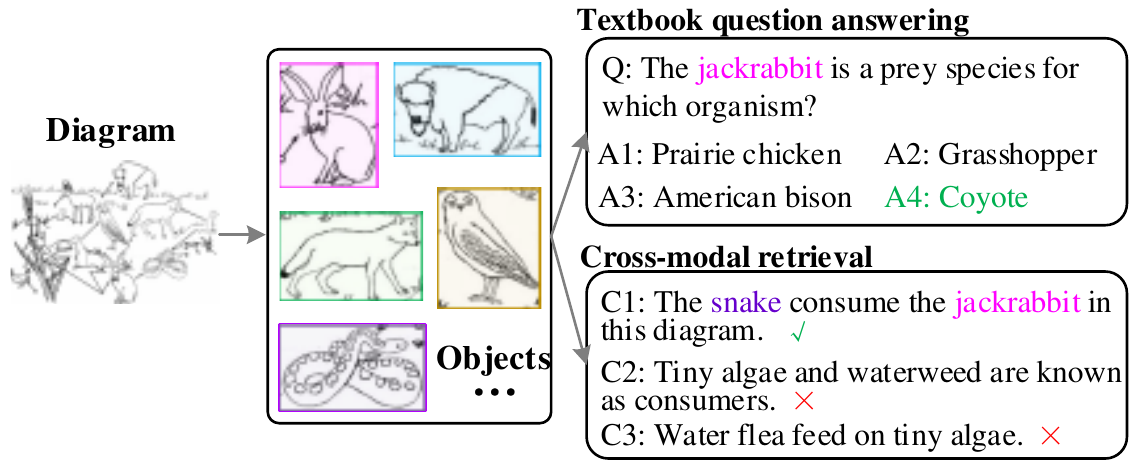}
	\caption{Efficient diagram object detectors can assist textbook question answering. Q, A, and C represent question text, candidate answer, and caption respectively.}
	\label{fig1}
\end{figure}

\begin{figure*}[t]
	\centering
	\includegraphics[width=0.85\textwidth]{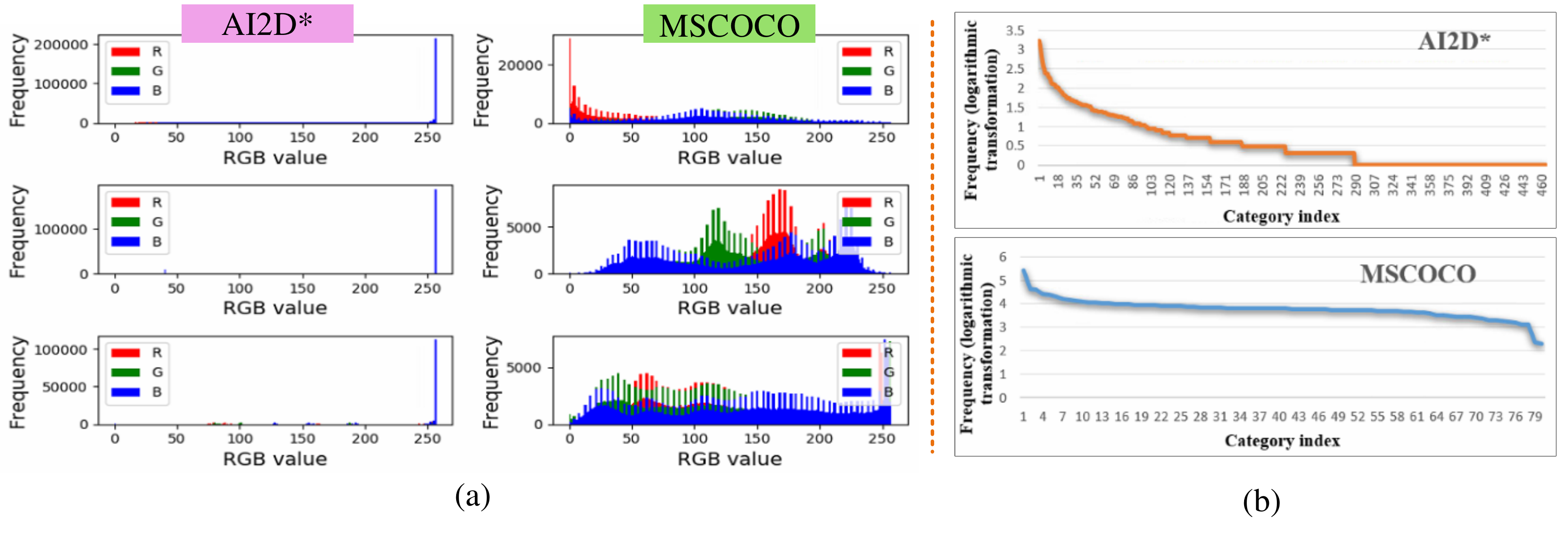}
	\caption{Comparative analyses of characteristics between diagram dataset AI2D* and natural image dataset MSCOCO.}
	\label{fig2}
\end{figure*}

Most existing detectors are designed for natural images of variant objects, while the research on diagram object detection is still blank. Diagram is a special kind of image, which usually consists of simple lines and color blocks, and exists in many fields such as pedagogy and architecture \cite{hu2021fs}. Diagram object detection is a key step in many applications as shown in Figure~\ref{fig1}. On this basis, it plays an important role in smart education and so on. Taking textbook question answering \cite{kembhavi2017you,he2021textbook} as an example, given the diagram and question text, diagram object detector outputs the locations and categories of objects in the diagram. And then, these objects interact with the question text multimodally to facilitate the answer of the question. However, detectors for natural images cannot be directly applied to diagram object detection. We adopt some mainstream detectors to conduct experiments on natural images and diagrams, respectively. Taking the recent SAM-DETR \cite{zhang2022accelerating} model as an example, the average precision of this model on natural images is as high as 39\%, while the precision on diagrams drops to about 15\%. See Section \ref{section4} for more analyses of experimental results. The reason is that the diagram has two characteristics different from natural image. \textbf{On the one hand, the visual features of diagram are sparser than those of natural image.} As shown in Figure~\ref{fig2} (a), the frequency distribution histograms are drawn corresponding to the RGB values for all pixels from the diagrams in AI2D* dataset and the natural images in MSCOCO \cite{lin2014microsoft}. We can see that the RGB value distribution of the natural image is more balanced than
that of the diagram, and the distribution of three RGB color components in the diagram is extremely uneven, which is concentrated around 255. This phenomenon illustrates the presence of a large amount of white in diagrams and the rest of the color information is scarce. White usually represents the background, which contains almost useless information. Therefore, there are a large number of white backgrounds in the diagrams, resulting in sparser visual features and fewer pixels occupied by foreground content compared with the natural images. \textbf{On the other hand, the ratio of low-frequency object categories is larger in the diagram.} In Figure~\ref{fig2} (b), the orange line depicts the long-tail distribution of object category in diagram dataset AI2D*. For MSCOCO, there is little difference in the frequency of all object categories. In summary, the existing detectors are not suitable for the task of diagram object detection.

How can humans efficiently identify the objects? According to the process of human perception \cite{wagemans2012century,pomerantz1977perception}, the human visual system tends to perceive patches in an image that are similar, close or connected without abrupt directional changes as a perceptual whole object. For example, in a jigsaw puzzle, humans consciously splice two patches with similar colors and close positions into a whole, and the spliced object has a smooth and continuous contour. Gestalt perception contains a series of laws to explain human perception, such as laws of similarity, proximity, closeness, smoothness, symmetry and so on. The diagram is drawn by experts and the object recognition process conforms to the gestalt perception theory \cite{wertheimer1922untersuchungen,horhan2021gestalt,desolneux2004gestalt}. Among them, similarity, proximity and smoothness laws play an important role in recognizing objects. 

Inspired by this, we propose a \textbf{G}estalt-\textbf{P}erception \textbf{TR}ansformer model for diagram object detection (GPTR). GPTR is based on the transformer encoder-decoder architecture, and the main module is the gestalt-perception graph named GPG that is constructed during encoding. Gestalt laws are used as prior knowledge to guide the aggregation of diagram patches to form reasonable objects, without relying on large amounts of annotations. The way of dividing the diagram into patches is the same as that of dividing the image into patches in \cite{dosovitskiy2020image}. Specifically, GPG is composed of diagram patches as nodes and the relationships between patches as edges. Node features in GPG are obtained by three gestalt-visual branches, namely color branch, position branch and edge branch. Edge weights of the graph are adaptively learned by the laws of color similarity, position proximity, and contour smoothness. The decoder of GPTR decodes the object queries in parallel and predicts the final location and classification results. Our main contributions are summarized as follows:

\begin{itemize}
	\item As far as we know, we put forward the diagram object detection task for the first time. Due to the problems of sparser visual features and more low-frequency objects of diagrams than those of natural images, we propose a novel gestalt-perception model to complete this task. The model is based on transformer architecture, and it can simulate the process of human visual perception to learn better features for diagram object detection. 
	\item We build a gestalt-perception graph, in which the adaptive learning strategy of gestalt-visual branches simulates humans to combine the diagram patches into more meaningful objects in accordance with the gestalt laws. In addition, we adopt the multi-scale attention mechanism to produce better query initialization.
	\item We conduct experiments on a diagram dataset AI2D* and a benchmark MSCOCO of natural images to verify the effectiveness of GPTR. The experimental results show that our model achieves the best results in the diagram object detection task, and also obtains comparable results over the competitors in natural images.
\end{itemize}

\begin{figure*}[t]
	\centering
	\includegraphics[width=0.79\textwidth]{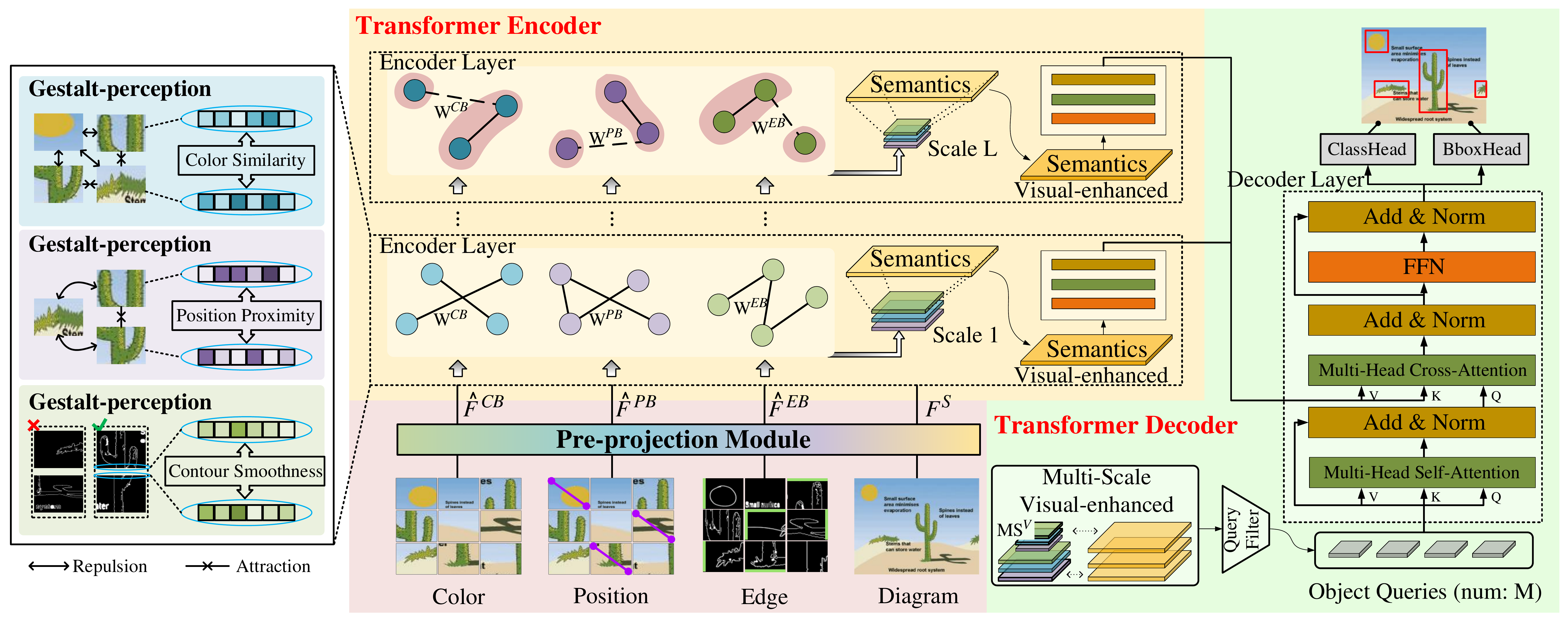}
	\caption{The overview architecture of our proposed gestalt-perception transformer model (GPTR). At each transformer encoder layer, we construct gestalt-perception graph to aggregate node features according to different laws. The whole process is stacked for $L$ layers and the decoder layer is used to decode $M$ object queries in parallel to predict the location and category of objects.}
	\label{fig3}
\end{figure*}

\section{Related Work}
This section mainly introduces DETR-series detection models and the gestalt perception theory.

\paragraph{DETR-Series Detection.}
DETR \cite{carion2020end} is the first end-to-end transformer-based detection model, which effectively removes the need for many hand-designed components. Subsequently, there are some improved models. Deform-DETR \cite{zhu2020deformable} designs a deformable attention module, which attends to a small set of sampling locations for prominent key elements out of all the feature map pixels. ConditionDETR \cite{meng2021conditional} learns a conditional spatial query from decoder embedding, while DAB-DETR \cite{liu2021dab} presents a novel query formulation using dynamic anchor boxes for DETR. SMCA-DETR constrains co-attention responses to be high near initially estimated bounding box locations. SAM-DETR interpretes its cross-attention as a matching and distillation process and semantically aligns object queries with encoded image features to facilitate their matching. The above models adopt convolutional network, which cannot effectively represent diagrams because of the sparse visual features. 

\paragraph{Gestalt Perception Theory.}
According to the process of human perception \cite{pomerantz1977perception}, characterized by the laws of similarity, proximity, and continuity, the human visual system tends to perceive objects that are similar, close or connected without abrupt directional changes as a perceptual whole. For example, GLGOV \cite{yan2018unsupervised} is guided by the gestalt laws of perception for image saliency detection with a bottom-up mechanism. Inspired by the gestalt laws of feature grouping, we propose a gestalt-perception model. Similarity, proximity and smoothness laws are considered in our work.

\begin{figure*}[t]
	\centering
	\includegraphics[width=0.76\textwidth]{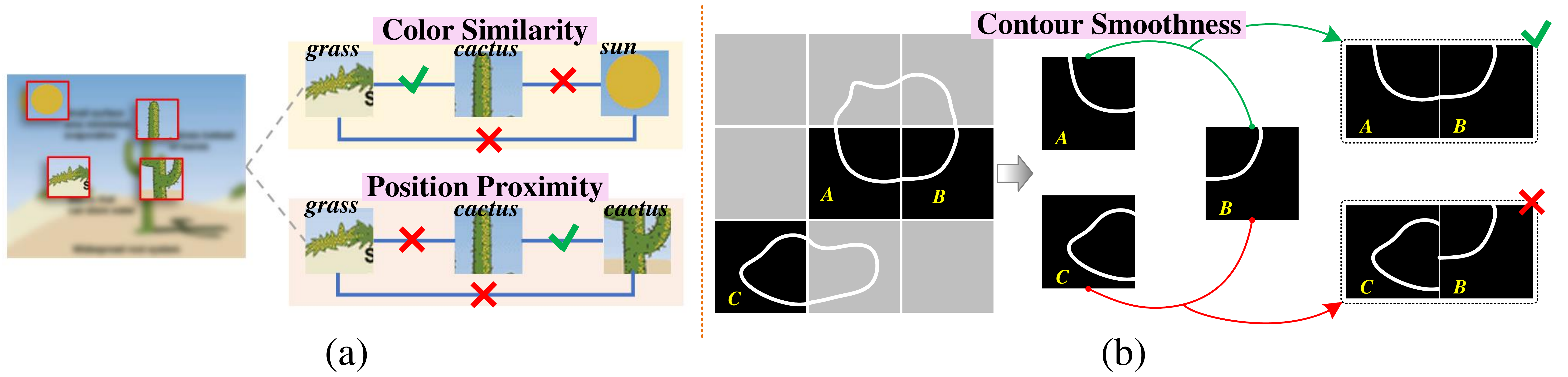}
	\caption{Example of color similarity, position proximity, and contour smoothness. \textit{A}, \textit{B} and \textit{C} in (b) indicate three randomly sampled patches in the diagram, and the white curve refers to the contour existing in the patch. {\faCheck} and {\faTimes} respectively indicate whether the gestalt law is conformed to.}
	\label{fig4}
\end{figure*}

\section{The GPTR Model}
\label{section3}
The overall architecture of GPTR is depicted in Figure~\ref{fig3}. It follows the encoder-decoder transformer and mainly contains three components: 1) the pre-projection module maps the patch features from different visual branches to the same dimensional space to initialize GPTR model; 2) the transformer encoder is built by gestalt-perception graph, which aims to model the relationships between the diagram patches and group these patches into objects via gestalt laws, so that the meaningful features can be better processed by the detector; 3) the transformer decoder transforms the object queries that represented by learnable positional embeddings into an output embedding and makes the final prediction with a feed-forward neural network (FFN). GPTR model is optimized with classification loss and box regression loss that are same as DETR \cite{carion2020end}. These three components are detailed in the following subsections.

\subsection{Pre-Projection Module}

The global features generated by only convolutional backbone network can not effectively represent diagrams because of the sparse visual features of diagrams. In order to make up for this limitation, we divide the diagram into local-level patches and let the GPTR model focus on the details of the diagram. Specifically, given an initial diagram $d\in{\mathbb{R}^{H_{0}\times W_{0}\times 3}}$, we reshape it into a set of patches $d^{P}=\{d^{P}_{i}\in{\mathbb{R}^{\frac{H_{0}}{\sqrt{N}}\times \frac{W_{0}}{\sqrt{N}}\times 3}},i=1,\cdots,N\}$ that is same as \cite{dosovitskiy2020image}. $(H_{0},W_{0})$ is the resolution of diagram $d$ and $3$ means three color channels. $N$ is the total number of patches in a diagram. Then, the pre-projection module learns patch features by feeding $d^P$ into three different gestalt-visual branches. In addition, this module adopts another MLP layer to project the backbone feature of the diagram into the $d$-dimension signed as $F^S$.

\textbf{Color Branch} (CB) maps per patch $d^{P}_{i}$ into a 9-dim color feature $f^{CB}_{i}$, and all the features consist of the feature set $F^{CB}=\{f^{CB}_{i}\in{\mathbb{R}^{1\times 9}},i=1,\cdots,N\}$. Specifically, $f^{CB}_{i}$ is concatenated by three central moments \cite{stricker1995similarity} of each color channel. These three moments represent mean feature, variance feature, and skewness feature of color distribution, respectively.

\textbf{Position Branch} (PB) outputs a position feature $f^{PB}_{i}$ for each patch, and $F^{PB}=\{f^{PB}_{i}\in{\mathbb{R}^{1\times 4}},i=1,\cdots,N\}$. $f^{PB}_{i}$ is composed of the coordinates of the top left corner $(x_{i0},y_{i0})$ and the bottom right corner $(x_{i1},y_{i1})$ of patch $d^{P}_{i}$. 

\textbf{Edge Branch} (EB) represents each patch $d^{P}_{i}$ as the pixel values of top, bottom, left, and right edges. We use Canny algorithm \cite{canny1986computational} converting 3-channel patch into 1-channel contour map. Then, the pixel values of the four edges of each contour map are concatenated as the edge feature for per patch. $F^{EB}=\{f^{EB}_{i}\in{\mathbb{R}^{1\times(2 \times \frac{W_0}{\sqrt{N}} + 2 \times \frac{H_0}{\sqrt{N}})}},i=1,\cdots,N\}$ denotes a set of edge features for each diagram. We use $F^{EB}_{(t;b)}$ and $F^{EB}_{(l;r)}$ to distinguish the top and bottom edge features from the left and right edge features.

In order to facilitate the construction and updating of the gestalt-perception graph, the pre-projection module adopts three kinds of MLP layers mapping low-dimensional visual features into high-dimensional ones as shown in follows. $d_c$, $d_p$ and $d_e$ are mapping dimensions for color, position and edge features. $||$ denotes the concatenating operator in \eqref{eq:2}.

\begin{equation}
\begin{split}
\hat{F}^{CB} = \mbox{MLP}^{CB}(F^{CB}), \hat{F}^{CB} \in \mathbb{R}^{N\times d_c},\\
\hat{F}^{PB} = \mbox{MLP}^{PB}(F^{PB}), \hat{F}^{PB} \in \mathbb{R}^{N\times d_p},
\end{split}
\label{eq:1}
\end{equation}
\begin{equation}
\begin{split}
\hat{F}^{EB} = \mbox{MLP}^{EB}(F^{EB}_{(t;b)}) || \mbox{MLP}^{EB}(F^{EB}_{(l;r)}), \\ \hat{F}^{EB} \in \mathbb{R}^{N\times 4 \times d_e}.
\end{split}
\label{eq:2}
\end{equation}

\subsection{Gestalt-Perception Graph in Encoder}
According to the process of human perception, the human visual system tends to perceive similar, close, or connected patches as a perceptual whole object. As shown by \textbf{the color similarity and position proximity} in Figure~\ref{fig4} (a), for the \textit{cactus} patch and \textit{sun} patch, because their color features are quite different, the two patches may not belong to the same object according to the law of color similarity. The same green \textit{grass} patch and \textit{cactus} patch are consistent with color similarity, but they are far away in spatial and do not meet the law of position proximity. Consequently, they can not belong to the same object. On the contrary, two different \textit{cactus} patches that are both green and close to each other can be easily recognized as the same object. Taking \textbf{the contour smoothness} in Figure~\ref{fig4} (b) as an example, according to the left and right edges for each patch, the features of the left edge of patch \textit{B} and the right edge of patch \textit{A} are similar, which means that the contours in \textit{B} and \textit{A} can be connected into a smooth curve, and they are more likely belong to the same object. On the contrary, the features of the left edge of patch \textit{B} and the right edge of patch \textit{C} are quite different, that is, if \textit{B} and \textit{C} are spliced into one object, it does not meet the human perception of the contour smoothness law. 

These gestalt laws, as a kind of priori knowledge, guide human to effectively identify the objects in the diagrams without relying on a large annotated dataset. Therefore, gestalt-perception based method can learn good representations for low-frequency objects. Inspired by this, GPTR designs a gestalt-perception graph (GPG) and it is composed of diagram patches as nodes and the relationships between patches as edges. GPG consists of three subgraphs, in other words, similarity, proximity and smoothness are encoded by the edges on subgraphs $\mathcal{G}^{CB}$, $\mathcal{G}^{PB}$ and $\mathcal{G}^{EB}$, respectively.

\paragraph{Color Similarity.} $\mathcal{G}^{CB}=(\mathcal{N}^{CB},\mathcal{E}^{CB})$ is a subgraph for modeling color similarity between patches. $\mathcal{N}^{CB}=\hat{F}^{CB}\in \mathbb{R}^{N\times d_c}$ indicates $N$ nodes, each node is a $d_c$-dimension color feature for one patch. $\mathcal{E}^{CB}\subseteq \mathcal{N}^{CB} \times \mathcal{N}^{CB}$ represents the color similarity between nodes. Specifically, given two node features $\hat{F}^{CB}_i$ and $\hat{F}^{CB}_j$, the weight of $\mathcal{E}^{CB}_{ij}$ is given by \eqref{eq:3}, where $\mbox{sim}(\cdot)$ is a cosine similarity function and $i,j=\{1,\cdots,N\}$.

\begin{equation}
W^{CB}_{ij}=\mbox{sim}(\hat{F}^{CB}_i,\hat{F}^{CB}_j).
\label{eq:3}
\end{equation}

\paragraph{Position Proximity.} In order to measure the proximity of spatial positions, $\mathcal{G}^{PB}=(\mathcal{N}^{PB},\mathcal{E}^{PB})$ is formulated to learn the positional relation between two patches. Concretely, $\mathcal{N}^{PB}$ indicates the nodes set with position feature $\hat{F}^{PB} \in \mathbb{R}^{N\times d_p}$, and $\mathcal{E}^{PB}$ is denoted as position proximity between each pair of nodes in $\mathcal{N}^{PB}$. The weight of $\mathcal{E}^{PB}_{ij}$ is shown in \eqref{eq:4}. The parameter $\delta$ is fixed as 0.1 and $i,j=\{1,\cdots,N\}$.

\begin{equation}
W^{PB}_{ij}=\mbox{exp}(-\frac{\sqrt{\sum_{t=1}^{d_p}(\hat{F}^{PB}_{it}-\hat{F}^{PB}_{jt})^2}}{\delta}).
\label{eq:4}
\end{equation}

\paragraph{Contour Smoothness.} The law of contour smoothness is one of the gestalt laws that states humans perceive objects as continuous in a smooth pattern, which means that object usually contains a smooth contour. In order to judge whether two patches may belong to the same object, $\mathcal{G}_{EB}$ is constructed to measure the feature consistency of the top, bottom, left and right edges between patches. Specifically, $\mathcal{G}_{EB}$ is defined as $\mathcal{G}_{EB}=(\mathcal{N}^{EB},\mathcal{E}^{EB})$. $\mathcal{N}^{EB}$ indicates the nodes with edge features $\hat{F}^{EB} \in \mathbb{R}^{N\times 4 \times d_e}$ and $\mathcal{E}^{EB}$ determines the possibility of splicing two patches. The weight of $\mathcal{E}^{EB}$ is computed as follows, where $i,j=\{1,\cdots,N\}$ and $\hat{F}^{EB_{b}}_i$ represents the bottom edge feature of patch $d_i^P$.

\begin{equation}
\sigma_{1}=\mbox{sim}(\hat{F}^{EB_{b}}_i,\hat{F}^{EB_{t}}_j);\sigma_{2}=\mbox{sim}(\hat{F}^{EB_{t}}_i,\hat{F}^{EB_{b}}_j),
\label{eq:6}
\end{equation}
\begin{equation}
\sigma_{3}=\mbox{sim}(\hat{F}^{EB_{l}}_i,\hat{F}^{EB_{r}}_j);\sigma_{4}=\mbox{sim}(\hat{F}^{EB_{r}}_i,\hat{F}^{EB_{l}}_j),
\label{eq:7}
\end{equation}
\begin{equation}
W^{EB}_{ij}=\mbox{max}\{\sigma_{1},\sigma_{2},\sigma_{3},\sigma_{4}\}.
\label{eq:8}
\end{equation}

\paragraph{GPG Grouping with an Assignment Matrix.} For aggregating patch features to obtain meaningful object features, we denote a learned assignment matrix \cite{ying2018hierarchical} at layer $l$ as $S^{(l)}\in \mathbb{R}^{N_{l}\times N_{l+1}}$, where $N_{l}$ is the number of nodes at layer $l$. It provides a soft assignment of each node at layer $l$ to layer $l+1$. Taking $\mathcal{G}_{CB}$ as an example, when $l=0$, $\hat{F}^{CB(l-1)}$ in \eqref{eq:9} denotes the output of pre-projection module. The node feature $\hat{F}^{CB(l)}$ at layer $l$ is computed by \eqref{eq:10}. The node update method of $\mathcal{G}_{PB}$ and $\mathcal{G}_{EB}$ is similar as that of $\mathcal{G}_{CB}$. GPG concatenates $\hat{F}^{CB(l)}$, $\hat{F}^{PB(l)}$ and $\hat{F}^{EB(l)}$, where $\alpha$, $\beta$ and $\gamma$ are three learnt adaptive weight coefficients. Then, a self-attention layer ($\mbox{SA}$) is applied to generate the final visual feature $F^{V(l)}$ as shown in \eqref{eq:11}. 

\begin{equation}
\tilde{F}^{CB(l-1)}=W^{CB(l-1)} \times \hat{F}^{CB(l-1)},
\label{eq:9}
\end{equation}
\begin{equation}
\hat{F}^{CB(l)}=S^{(l)^{\top}}\times \tilde{F}^{CB(l-1)},
\label{eq:10}
\end{equation}
\begin{equation}
F^{V(l)}=\mbox{SA}(\alpha \times \hat{F}^{CB(l)}||\beta \times \hat{F}^{PB(l)}||\gamma \times \hat{F}^{EB(l)}).
\label{eq:11}
\end{equation}

GPTR updates the diagram feature of visual enhancement through cross-attention strategy ($\mbox{CA}$), then a self-attention layer ($\mbox{SA}$) and a feed-forward layer ($\mbox{FFN}$) are stacked to form a transformer encoder layer. $F^S$ in \eqref{eq:12} and \eqref{eq:13} indicates the high-level semantic feature extracted from Convolution Neural Network (CNN).

\begin{equation}
\mbox{CA}(F^{S},F^{V(l)})=\mbox{softmax}(F^{S},F^{V(l)^{\top}}) \times F^{V(l)},
\label{eq:12}
\end{equation}
\begin{equation}
F_{\mbox{ENCODER}}^{(l)}=\mbox{FFN}(\mbox{SA}(\mbox{CA}(F^{S},F^{V(l)}) + F^S)).
\label{eq:13}
\end{equation}

\subsection{Multi-Scale Visual-Enhanced Decoder}
The decoder follows the standard architecture of ConditionDETR \cite{meng2021conditional}, transforming $M$ embeddings using multi-head self-attention and cross-attention mechanisms. 
Unlike ConditionDETR decoder, which receives zero set as initial queries, we consider the human visual perception. When recognizing objects in diagrams, humans follow the process of visual perception to identity variety of objects with different scales. Inspired by this, our GPTR designs a multi-scale attention mechanism named $\mbox{MSA}$, to acquire better initial query features.

Firstly, the output of each layer of $\mathcal{G}^{CB}$, $\mathcal{G}^{PB}$ and $\mathcal{G}^{EB}$ is taken as the multi-scale visual feature. Taking $\mathcal{G}^{CB}$ as an example, the multi-scale color feature is recorded as $\mbox{MS}^{CB}=[\hat{F}^{CB(1)},\cdots,\hat{F}^{CB(l)},\cdots,\hat{F}^{CB(L)}]$. The score of color feature $\mbox{SCORE}^{CB}$ is computed by a single-layer $\mbox{MLP}$ as shown in \eqref{eq:15}, and top-$M$ color feature $\tilde{\mbox{MS}}^{CB}$ is selected according to the score. $\tilde{\mbox{MS}}^{PB}$ and $\tilde{\mbox{MS}}^{EB}$ are acquired in the same way as $\tilde{\mbox{MS}}^{CB}$. The final selected multi-scale visual feature is written as $\mbox{MS}^{V}=\tilde{\mbox{MS}}^{CB}+\tilde{\mbox{MS}}^{PB}+\tilde{\mbox{MS}}^{EB}$. In \eqref{eq:16}, GPTR first concatenates the $ \mbox{MS}^{V} $ and the output feature $ F_{\mbox{ENCODER}}^{(L)} $ of the $L$-layer encoder, and then obtains the enhanced features through the self-attention mechanism SA. $[:M]$ indicates that the first $M$ features are selected as the initial query representation $\mbox{QUERY}$.

\begin{equation}
\mbox{SCORE}^{CB}=\mbox{softmax}(\mbox{MLP}(||_{l=1}^{L}\hat{F}^{CB(l)})),
\label{eq:15}
\end{equation}
\begin{equation}
\mbox{QUERY}=\mbox{SA}(\mbox{MS}^{V}||F_{\mbox{ENCODER}}^{(L)})[:M].
\label{eq:16}
\end{equation}

\paragraph{Loss Function.} We follow DETR to find an optimal bipartite matching \cite{kuhn1955hungarian} between the predicted and ground-truth objects using the Hungarian algorithm, and then form the loss function for optimizing GPTR model. Focal loss \cite{lin2017focal} is used for classification and GIoU loss \cite{rezatofighi2019generalized} for box regression, both of which are the same as DETR.

\section{Experiments}
\label{section4}
\subsection{Datasets}
In this work, we evalute the baselines and our GPTR model both on the diagram and the natural image datasets.
 
\textbf{AI2D*} is composed of diagrams in the original AI2D dataset \cite{kembhavi2016diagram}, and the topic is grade school science. AI2D is mainly used to verify the question and answering task. We annotate it with more fine-grained details, including the spatial coordinates and category labels of objects in per diagram. The novel AI2D* dataset contains total 557 object categories and it is divided into a train set with 1,634 diagrams and a test set with 404 diagrams. 

\textbf{MSCOCO} \cite{lin2014microsoft} is a large-scale object detection dataset with 80 categories. It comprises 118,287 images for training and 5,000 images for testing.

\begin{table*}[t]
	\centering
	\begin{tabular}{l|cccc|cccccc|c}
		\hline
		Models & $L$ & $H$ & $BS$ & Epoch & AP & AP50 & AP75 & APS & APM & APL & params \\
		\hline
		CenterNet \cite{duan2019centernet} &/ &/ &16 &500 &8.6 &13.2 &9.9 &10.7 &13.4 &12.5 & 50.39M \\
		RetinaNet \cite{lin2017focal} &/ &/ &16 &100 &10.5 &16.3 &11.4 &6.0 &12.8 &14.9 & 29.86M\\ 
		DETR \cite{carion2020end} &4 &4 &16 &1000 &10.5 &18.3 &11.0 &6.6 &13.7 &13.9 &28.93M \\
		ConditionDETR \cite{meng2021conditional} &4 &4 &16 &100 &11.5 &18.5 &12.7 &10.4 &15.1 &15.2 &29.22M \\ 
		\textbf{GPTR(Ours)} &4 &4 &16 &100 &\textbf{14.1} &\textbf{23.0} &\textbf{15.6} &\textbf{12.2} &\textbf{18.4} &\textbf{18.9} &30.56M \\ \hline
		Deform-DETR \cite{zhu2020deformable} &6 &8 &8 &100 &11.8 &16.8 &14.2 &13.9 &15.8 &16.6 &35.11M \\
		DAB-DETR \cite{liu2021dab} &6 &8 &8 &100 &10.8 &17.1 &12.0 &14.9 &14.1 &14.6 &41.55M \\
		SMCA-DETR \cite{gao2021fast} &6 &8 &8 &300 &13.8 &21.7 &15.4 &10.5 &18.1 &18.4 &39.66M \\
		SAM-DETR \cite{zhang2022accelerating} &6 &8 &8 &200 &14.6 &21.7 &16.6 &10.9 &19.0 &18.5 &47.08M \\
		AnchorDETR \cite{wang2022anchor} &6 &8 &8 &120 &15.6 &23.5 &17.3 &14.8 &19.4 &20.5 &32.22M \\
		\textbf{GPTR(Ours)} &6 &8 &8 &120 &\textbf{16.1} &\textbf{24.6} &\textbf{18.4} &\textbf{15.3} &\textbf{21.1} &\textbf{21.5} &33.44M \\ \hline
		
	\end{tabular}
	\caption{The precision (\%) comparison on challenging AI2D* dataset for diagram object detection. $L$, $H$, and $BS$ represent the layer number of transformer encoder-decoder, the number of attention heads and batchsize respectively.}
	\label{table1}
\end{table*}

\begin{table*}[t]
	
	\centering
	\begin{tabular}{l|cccc|cccccc}
		\hline	
		Models & $L$ & $H$ & $BS$ & Epoch & AP & AP50 & AP75 & APS & APM & APL \\
		\hline
		CenterNet \cite{duan2019centernet} &/ &/ &32 &100 &20.1 &39.6 &16.6 &7.5 &22.8 &29.1 \\
		RetinaNet \cite{lin2017focal} &/ &/ &16 &100 &25.5 &42.4 &26.2 &10.7 &27.6 &38.0 \\
		DETR \cite{carion2020end} &4 &4 &16 &1000 &30.2 &49.7 &30.6 &10.1 &31.4 &47.2  \\
		ConditionDETR \cite{meng2021conditional} &4 &4 &16 &50 &31.9 &52.4 &32.6 &13.7 &34.0 &48.7 \\ 
		\textbf{GPTR(Ours)} &4 &4 &16 &50 &\textbf{32.1} &\textbf{52.4} &\textbf{33.2} &\textbf{13.8} &\textbf{34.7} &\textbf{48.8} \\ \hline 
		SMCA-DETR \cite{gao2021fast} &6 &8 &8 &50 &28.9 &50.3 &28.7 &9.9 &30.7 &46.9 \\
		SAM-DETR \cite{zhang2022accelerating} &6 &8 &8 &50 &39.0 &60.5 &40.8 &\textbf{19.7} &42.5 &58.0 \\
		\textbf{GPTR(Ours)} &6 &8 &8 &50 &37.2 &57.8 &38.2 &14.0 &41.6 &54.9 \\
		
		SAM-DETR+\textbf{GPG} &6 &8 &8 &50 &\textbf{39.3} &\textbf{61.5} &\textbf{41.2} &19.5 &\textbf{43.3} &\textbf{58.9} \\
		\hline
	\end{tabular}
	\caption{The precision (\%) comparison on benchmark MSCOCO dataset for natural image object detection.}
	\label{table2}
\end{table*}

\subsection{Experimental Settings}
\paragraph{GPTR Implementation.} Our architecture is almost the same with the DETR-like architecture and contains the CNN backbone, transformer encoder and transformer decoder. The main difference is that we introduce the details of gestalt-perception graph in transformer encoder. For the gestalt visual preprocessing, we resize all images of two datasets to 224$\times$224$\times$3 and each image is divided into 196 patches. The dimension of per patch feature is $d_{c}=d_{p}=d_{e}=$256. We set 50 and 100 object queries for AI2D* and MSCOCO datasets, respectively.

\paragraph{Training and Evaluation.} The learning rate is initially set to $10^{-4}$ and the AdamW optimizer is used in GPTR. The weight decay is set to be $10^{-4}$ and the dropout rate in transformer is 0.1. We use the standard COCO evaluation introduced in \cite{meng2021conditional}, and we also report the average precision (AP), and the AP scores at 0.50 (AP50), 0.75 (AP75) and for the small (APS), medium (APM), and large (APL) objects. For fair comparison, we adopt the same equipment and settings, such as the layer number $L$ for transformer encoder and decoder, and the number of attention heads $H$ inside the transformer's attentions, to rerun all the baseline models for several times, and then record the average results. All the models are trained and evaluated on NVIDIA Tesla V100 GPU.

\subsection{Performance Comparison}
\paragraph{Diagram Object Detection.} We conduct this experiment on the AI2D* dataset and the proposed GPTR achieves the best results compared with all the competitors. One can find in Table~\ref{table1} that DETR with 1,000 training epochs performs much worse than ConditionDETR with only 100 epochs. The performance of our GPTR is 1.8\% to 4.5\% higher than that of ConditionDETR in all AP scores. Compared with SMCA-DETR and SAM-DETR, GPTR achieves better results in all AP scores. Concretely, GPTR is 2.3\% and 1.5\% higher in AP than SMCA-DETR and SAM-DETR, respectively. Especially for small objects, GPTR has a gain of 4.8\% in APS than that of SMCA-DETR. In addition, the parameters of SAM-DETR and SMCA-DETR are much more than our GPTR model, and the training time of GPTR is only less than half that of SMCA-DETR. Also, our GPTR outperforms the recently proposed AnchorDETR model in all AP scores, especially 0.5\% higher in AP.

\paragraph{Natural Image Object Detection.} Although GPTR model is especially proposed for the diagram object detection, it can also be applied to the object detection in natural images. For natural images, the patches also meet three gestalt laws of color similarity, position proximity and contour smoothness. The performance of GPTR is verified on MSCOCO with natural images as shown in Table~\ref{table2}. One can find that our model achieves competitive results on this task. Specifically, DETR works better than the anchor-free models CenterNet and RetinaNet, but it converges more slowly. The ConditionDETR model is built on the DETR model, with higher AP scores and faster convergence. Compared with ConditionDETR, our GPTR still achieves the best performance under the same experimental settings, and GPTR is 8.3\% and 4.1\% higher than SMCA-DETR in AP and APS scores, respectively. In addition, the designed gestalt-perception graph (GPG) in transformer encoder can be flexibly added to SAM-DETR model, and the performance of SAM-DETR+\textbf{GPG} is improved in almost all the AP scores.

\paragraph{Diagrams vs. Natural Images.} From the experimental results in Table~\ref{table1} and Table~\ref{table2}, it can be seen that the GPTR outperforms almost all the competitors in AP scores. In particular, compared with natural image object detection, GPTR improves the performance of diagram object detection more significantly. In other words, the gestalt laws in GPTR are more effective for the representation of diagrams. The reason is that the visual features of diagrams are sparse, and there are many low-frequency object categories. As a kind of prior knowledge of human cognition, gestalt laws can effectively learn the visual features of diagrams without relying on a large amount of labeled dataset, and alleviate the limitations of learning low-frequency object representations.

\begin{table}[t]
	\centering
	\setlength{\tabcolsep}{0.9mm}{
		\begin{tabular}{l|ccc|cccccc}
			\hline	
			Model & AdaB & VQ & MSA & AP & APS & APM & APL\\
			\hline
			GVB-CPE &-  &-  &-  &11.6 &10.7 &15.5 &15.6\\
			GVB-CPE$^{a}$ & $\checkmark$ & & &11.9 &11.6 &16.0 &16.5 \\
			GVB-CPE$^{b}$ & $\checkmark$ & $\checkmark$ & &12.7 &11.7 &16.9 &18.0 \\
			GPTR & $\checkmark$ & $\checkmark$ & $\checkmark$ &\textbf{14.1} &\textbf{12.2} &\textbf{18.4} &\textbf{18.9} \\
			\hline
	\end{tabular}}
	\caption{Ablation studies on the AI2D* dataset. ``AdaB" indicates the adaptive combination of three gestalt-visual branches. ``VQ" represents the visual-guided query initialization. ``MSA" means adding multi-scale attention mechanism to GVB-CPE$^{b}$ to generate better query representations.}
	\label{table3}
\end{table}
\begin{table}[t]
	\centering
	\setlength{\tabcolsep}{1.4mm}{
		\begin{tabular}{l|ccc|cccccc}
			\hline	
			Model & CB & PB & EB & AP & APS & APM & APL\\
			\hline
			-ALL &-  &-  &-  &11.4 &10.3 &15.1 &15.2\\
			GVB-C & $\checkmark$ & & &11.0 &8.6 &14.3 &14.6\\
			GVB-P & & $\checkmark$ & &11.2 &7.4 &14.9 &14.8\\
			GVB-E & & & $\checkmark$ &10.6 &9.4 &13.4 &14.0 \\ \hline
			GVB-CPE & $\checkmark$ & $\checkmark$ & $\checkmark$ &\textbf{11.6} &\textbf{10.7} &\textbf{15.5} &\textbf{15.6}\\
			\hline
	\end{tabular}}
	\caption{Ablation studies on the AI2D* dataset. ``CB", ``PB" and ``EB" represent color branch, position branch, and edge branch respectively. ``-All" refers the model that only uses CNN backbone for extracting diagram features.}
	\label{table4}
\end{table}

\subsection{Ablation Studies}
The performance of GPTR in diagram object detection is mainly improved in three aspects. They are gestalt-visual branches in GPG module, visual-guided initialization for decoder queries, and multi-scale visual enhancement strategy. To demonstrate the effectiveness of these aspects, we study the ablation models and the differences between these versions are shown in Table~\ref{table3}. 1) GVB-CPE represents the combination of three branches as visual features, and the combination mode is direct concatenate. 2) GVB-CPE$^{a}$ adopts adaptive learning method to combine the three branches on the basis of GVB-CPE. 3) GVB-CPE$^{b}$ adopts visual-guided initialization mechanism, and our GPTR model adds the multi-scale mechanism to GVB-CPE$^{b}$.

The experimental results are shown in Table~\ref{table3}. One can find that: 1) after using the adaptive method, GVB-CPE$^{a}$ has significantly improved the APS score compared with GVB-CPE model. 2) Compared with taking zero set as initial queries in GVB-CPE$^{a}$, the visual-guided initialization mechanism of GVB-CPE$^{b}$ has improved performance in all AP scores, especially 1.5\% higher in APL. 5) Our GPTR, which adds multi-scale attention mechanism to GVB-CPE$^{b}$, has achieved the best results in all AP scores.

Since three gestalt-visual branches are included in the GPTR model, we also analyze the effects of different visual branches. As shown in Table~\ref{table4}, 1) GVB-C, GVB-P and GVB-E indicate that only the gestalt-visual branches of color, position and edge, respectively. 2) GVB-CPE is the same model in Table~\ref{table3}.	When three gestalt-visual branches are used separately, the performance of GVB-C, GVB-P and GVB-E is affected by the law-bias compared with -ALL model. 2) GVB-CPE has improved in most AP scores that compared with all the ablation models.

\subsection{Qualitative Results}
\begin{figure}[t]
	\centering
	\includegraphics[width=0.35\textwidth]{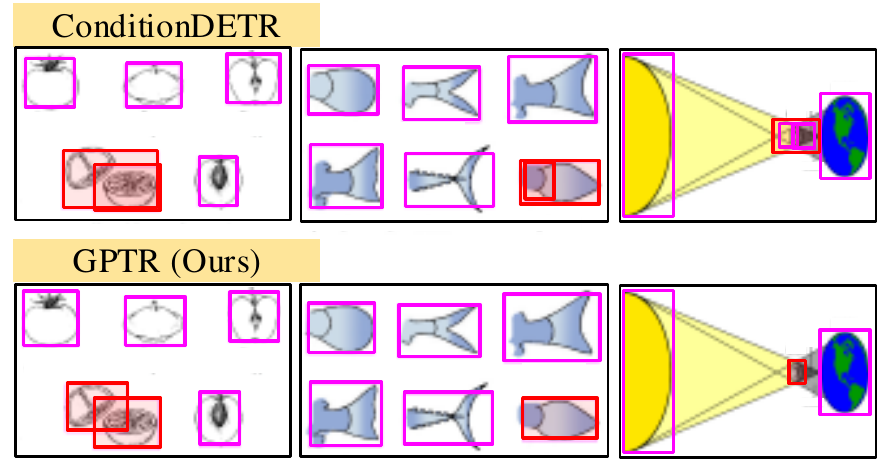}
	\caption{Qualitative results of ConditionDETR and our GPTR. We use the red bounding boxes to highlight the differences in the detection results between these two models.}
	\label{fig5}
\end{figure}

\paragraph{Visualization of Detection Results.} The detection results for ConditionDETR and GPTR are shown in Figure~\ref{fig5}. For the first case, when two \textit{orange} are close to each other in a diagram, ConditionDETR confuses them as a whole object, while GPTR accurately locates them respectively. For the second case, the \textit{fishtail} in the bottom right corner is composed of two polygons. ConditionDETR recognizes it as two independent objects, while GPTR accurately recognizes it as a whole \textit{fishtail}. For the third case, the \textit{moon}, \textit{earth} and \textit{light} in this diagram are close in space, and the light as the background affects the recognition of the foreground objects \textit{moon} and \textit{earth} by the ConditionDETR. On the contrary, GPTR effectively separates the foreground and background, and then accurately locate the foreground objects.

\begin{figure}[t]
	\centering
	\includegraphics[width=0.45\textwidth]{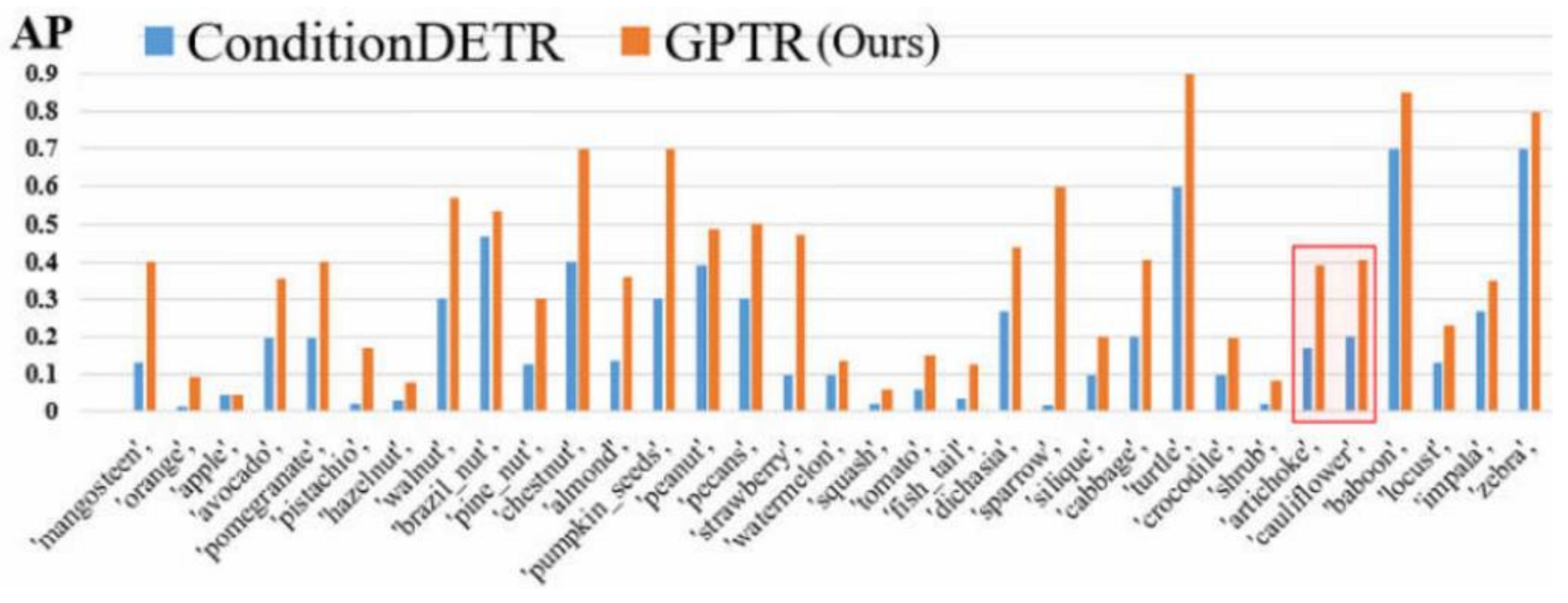}
	\caption{The AP score of low-frequency category for ConditionDETR and our GPTR.}
	\label{fig6}
\end{figure}

\paragraph{Low-Frequency Objects in AI2D* Dataset.} Figure~\ref{fig6} shows the AP score of the ConditionDETR and GPTR models on low-frequency objects, respectively. For the convenience of visualization, the abscissa represents several object categories selected with a frequency of no more than 10 times, and the ordinate indicates the AP score. It can be seen that the performance of GPTR is better than that of the ConditionDETR for low-frequency object categories. Especially for some categories that only appear once, such as \textit{artichoke} and \textit{cauliflower} in the red box, the performance of GPTR is about 20\% higher than that of ConditionDETR.


\section{Conclusion}
In this paper, we propose a gestalt-perception transformer model (GPTR) for the novel diagram object detection. The gestalt laws, as a kind of priori knowledge, guide human to identify the objects without relying on a large dataset. For the sparse visual features and low-frequency objects of diagrams, GPTR constructs a gestalt-perception graph and these laws are encoded by the graph edges. During updating, the designed adaptive learning strategy effectively combine the laws of similarity, proximity and smoothness to group the diagram patches to objects. In addition, we adopt the multi-scale mechanism based on the visual features to produce better queries. We have demonstrated the effectiveness of GPTR in diagram object detection by achieving significant performance improvements. However, there are still limitations in the application of gestalt laws in this work. For example, GPTR only uses three laws. How to mine other laws for diagram representation, and how multiple laws work together will be the future research works.

\section*{Acknowledgments}
This work was supported by National Key Research and Development Program of China (2020AAA0108800), National Natural Science Foundation of China (62137002, 61937001, 62192781, 62176209, 62176207, 62106190, and 62250009), Innovative Research Group of the National Natural Science Foundation of China (61721002), Innovation Research Team of Ministry of Education (IRT$\_$17R86), Consulting research project of Chinese academy of engineering ``The Online and Offline Mixed Educational Service System for `The Belt and Road' Training in MOOC China", ``LENOVO-XJTU" Intelligent Industry Joint Laboratory Project, CCF-Lenovo Blue Ocean Research Fund, Project of China Knowledge Centre for Engineering Science and Technology, Foundation of Key National Defense Science and Technology Laboratory (6142101210201).

\bibliography{aaai23}

\end{document}